\documentclass[10pt]{article} 

\usepackage[preprint]{tmlr}

\usepackage{amsmath,amsfonts,bm}

\def\eqref#1{equation~\ref{#1}}

\def\1{\bm{1}}

\DeclareMathAlphabet{\mathsfit}{\encodingdefault}{\sfdefault}{m}{sl}
\SetMathAlphabet{\mathsfit}{bold}{\encodingdefault}{\sfdefault}{bx}{n}

\usepackage{hyperref}
\usepackage{url}
\usepackage{graphicx}
\usepackage{booktabs}
\usepackage[table]{xcolor}

\usepackage{enumitem}
\usepackage{pif ont}    
\usepackage{multirow}
\usepackage{array}
\usepackage{makecell}
\usepackage{caption}
\usepackage{float}
\usepackage{placeins}
\usepackage{pifont}
\usepackage{xspace}
\usepackage{amsmath}
\usepackage{amssymb}
\usepackage{wrapfig}
\usepackage{adjustbox}
\usepackage{needspace}
\usepackage{placeins}
\usepackage{threeparttable}
\usepackage{graphicx} 
\usepackage{booktabs}
\usepackage{tabularx}

\newcolumntype{Y}{>{\centering\arraybackslash}X}

\def\onedot{\ifx\@let@token.\else.\null\fi\xspace}
\newcommand{\cmark}{\ding{51}} 
\newcommand{\xmark}{\ding{55}} 

\definecolor{my_blue}{HTML}{d3eaf2}
\definecolor{Green}{rgb}{0.85882353, 0.90980392, 0.84705882}
\definecolor{rose}{rgb}{0.60392157, 0.53333333, 0.43921569}
\definecolor{dred}{rgb}{0.7254902, 0.09803922, 0.10588235}

\def\eg{\emph{e.g}\onedot} 
\def\ie{\emph{i.e}\onedot} 

\usepackage{graphicx}
\usepackage{xcolor}
\usepackage{array}

\usepackage{graphicx}
\usepackage{multirow}
\usepackage[table]{xcolor}
\usepackage{subcaption}

\title{DIFFCZSL: Compositional Zero-Shot Learning Regularized by Diffusion Representations}

\author{\name Hangyu Tian,
      \name Zhenqi He\textsuperscript{\dag},
      \name Yanghao Wang\textsuperscript{\dag},
      \name Long Chen\textsuperscript{\ddag} \\
      \addr The Hong Kong University of Science and Technology}

\footnotetext{\textsuperscript{\dag} Project Leader.}
\footnotetext{\textsuperscript{\ddag} Corresponding Author. Email: longchen@ust.hk}

\def\month{MM}  
\def\year{YYYY} 
\def\openreview{\url{https://openreview.net/forum?id=XXXX}} 

\begin{document}

\maketitle

\begin{abstract}
Compositional Zero-Shot Learning (CZSL) aims to recognize unseen attribute-object compositions by leveraging knowledge of primitive concepts learned from seen compositions. Although recent works achieve impressive performance in CZSL by leveraging large vision-language models, they primarily rely on discriminative representations that may not explicitly preserve the structured relationships between primitive concepts and their compositions. Motivated by the recent success of diffusion-based classifiers and their competitive performance relative to discriminative models, we investigate whether intermediate diffusion representations can provide complementary cues for CZSL. To this end, we propose DIFFCZSL, a diffusion-augmented framework that injects generative priors from pre-trained diffusion models into CLIP-based CZSL pipelines. We extract intermediate diffusion representations and project them into the CLIP embedding space to provide auxiliary supervision on both image and text modalities. Through contrastive alignment between CLIP embeddings and diffusion features during training, our method encourages the embedding geometry toward richer composition-aware semantics, while introducing no additional cost at inference time. Extensive experiments on three public CZSL benchmarks demonstrate consistent improvements over strong CLIP-based baselines under both closed-world and open-world settings. Our results highlight the complementary strengths of generative diffusion representations and discriminative vision-language models for compositional generalization.
\end{abstract}

\section{Introduction}

\begin{figure}[t]
    \centering
    \includegraphics[width=\linewidth]{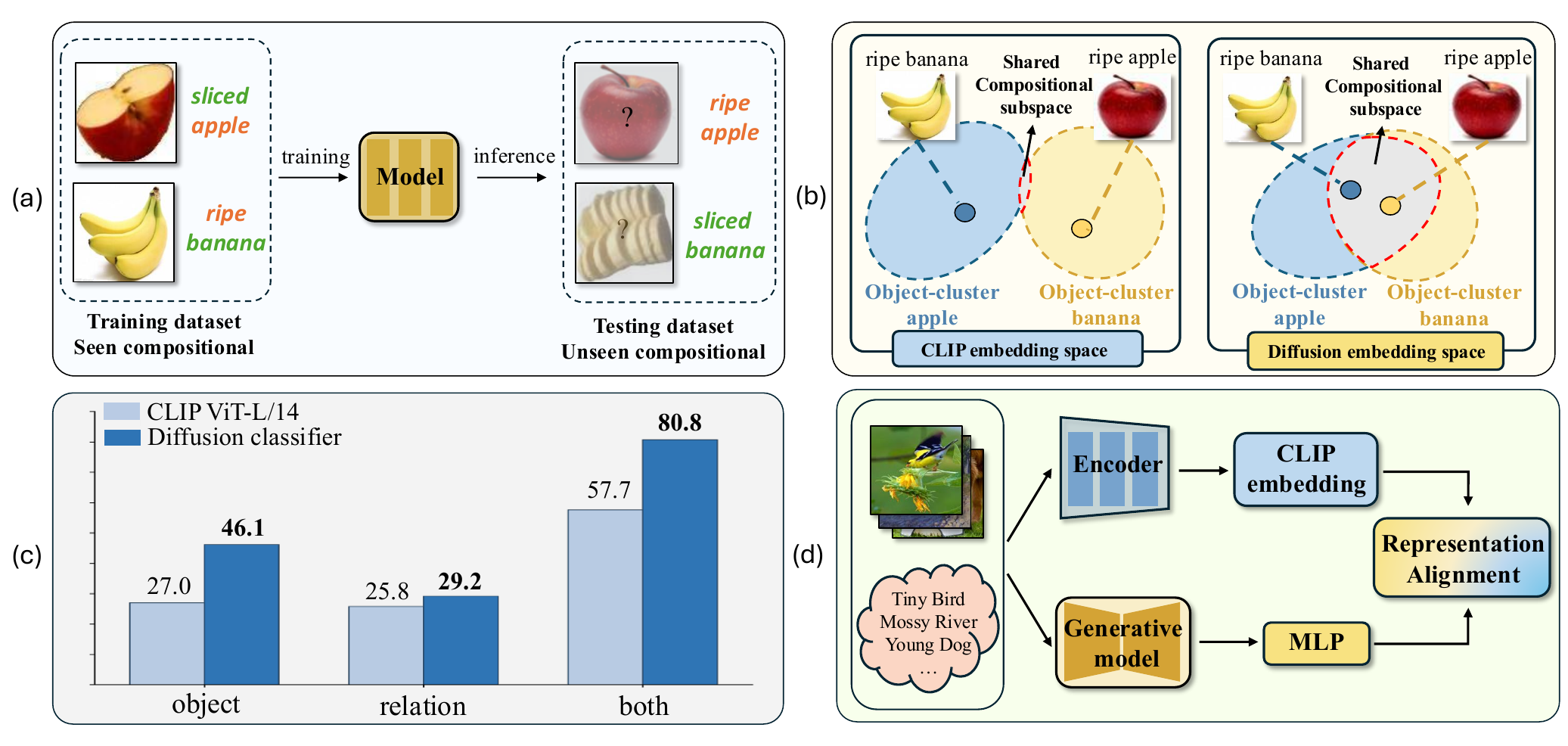}
    \caption{(a) CZSL aims to recognize unseen attribute-object compositions by learning from seen compositions. (b) Comparison between CLIP and diffusion embedding spaces. CLIP tends to form object-centric clusters, resulting in limited shared compositional structure across different objects. Diffusion representations exhibit more consistent compositional patterns, leading to a more structured and shared compositional subspace. (c) Prior evidence from the Winoground compositional reasoning benchmark, adapted from~\citep{li2023your}. Diffusion-based representations substantially outperform CLIP on object, relation, and joint compositional reasoning, suggesting that diffusion models capture compositional semantics more effectively than contrastive representations. (d) Overview of our approach: We propose to regularize the CLIP embeddings with diffusion features through representation alignment for better compositional generalization.}
    \label{fig:intro}
\end{figure}
Compositional Zero-Shot learning (CZSL) studies the problem of recognizing unseen attribute–object compositions at test time, where primitive concepts (\ie, attributes and objects) are observed in other compositions during training. As illustrated in Fig.~\ref{fig:intro}(a), models are trained on seen attribute–object pairs and evaluated on unseen compositions constructed from the same primitives. The central challenge is therefore to learn transferable representations of primitive concepts that can be effectively recomposed to generalize beyond the observed combinations. 
Motivated by this compositional principle, early CZSL approaches primarily rely on supervised visual learning to capture reusable attribute and object semantics. These methods learn pair-level classifiers, explicitly decompose images into primitive representations, or model their interactions for recognizing unseen compositions~\citep{purushwalkam2019task,atzmon2020causal,huynh2020compositional,karthik2022kg,li2022siamese,misra2017red}.
However, because such representations are learned largely from limited task-specific supervision, they provide only sparse coverage of the combinatorial attribute--object space, making them prone to biases toward seen compositions and limiting their ability to transfer primitive knowledge to unseen attribute--object combinations.

Recent CZSL methods build on large-scale vision--language models (VLMs) such as CLIP~\citep{radford2021learning}, whose pre-trained image and text encoders define a shared cross-modal representation space. Rather than learning a new compositional space from scratch, these methods formulate CZSL as matching CLIP-derived image embeddings against text embeddings of candidate attribute--object prompts. They mainly differ in how this pre-trained space is adapted. On the text side, prompt-level composition modeling introduces learnable context, attribute, or object tokens to bind primitives within compositional prompts; for instance, CSP~\citep{csp} concatenates learnable attribute and object tokens under a fixed prefix template (\eg, ``a photo of''), while subsequent works~\citep{huang2024troika,lu2023decomposed} introduce branch-specific prompts or representations to model the distinct semantic roles of attributes and objects, enabling more structured compositional modeling. On the image side, existing methods enhance visual representations through primitive disentanglement, attribute--object interaction modeling, or lightweight adapters for parameter-efficient fine-tuning, yielding more composition-aware visual features~\citep{PLID,huang2024troika,wu2025logiczsl,cams}.
Despite these architectural differences, existing methods still perform compositional recognition within CLIP’s contrastively pretrained embedding space. Their generalization is therefore ultimately constrained by the compositional structure encoded in this space.

Although CLIP-based approaches have substantially advanced CZSL, their reliance on contrastively pre-trained representations leaves two limitations that are particularly relevant to compositional generalization.
\textbf{(1) Limited intra-composition interaction modeling.} CLIP may match an image with a compositional prompt by relying on separate evidence for the attribute and the object, without explicitly modeling how the attribute modifies the object. For example, recognizing ``ripe banana'' requires not only identifying ripeness and banana identity, but also understanding how ripeness is visually manifested on a banana. Since the semantics of an attribute can vary with the object it modifies, independently matching primitive cues may provide an incomplete representation of the composition as a whole. This limitation is consistent with prior observations that vision--language models can exhibit bag-of-words behavior and limited sensitivity to compositional structure~\citep{yuksekgonul2022and}.
\textbf{(2) Limited inter-composition relational consistency.} Contrastive objectives encourage observed compositions to be distinguishable, but do not explicitly enforce that the same semantic change induces a consistent transformation across different object contexts. 
As illustrated in Fig.~\ref{fig:intro}(b), such representations may become dominated by object-specific clusters, with limited compositional structure shared across them. This limitation is especially consequential in CZSL, where unseen attribute--object pairs receive no direct supervision. Reliable compositional generalization therefore requires structure at two complementary levels: capturing attribute--object interactions within individual compositions and preserving transferable relations across different compositions. Consequently, discriminability on seen pairs alone may be insufficient for recognizing unseen compositions.

Inspired by the recent success of diffusion models in discriminative tasks and compositional reasoning, we revisit whether representations learned from generative objectives can benefit compositional zero-shot learning. As illustrated in Fig.~\ref{fig:intro}(c), recent studies~\citep{li2023your,krojer2023diffusion} have shown that diffusion models exhibit substantially stronger compositional reasoning than contrastive vision-language models on challenging benchmarks such as Winoground~\citep{thrush2022winoground}. Unlike CLIP's contrastive learning, which primarily emphasizes global image-text alignment and instance-level discrimination, text-conditioned diffusion models learn to recover visual content from noisy inputs through denoising~\citep{ho2020denoising,dhariwal2021diffusion,rombach2022high}. Such a generative objective requires the model to capture not only which concepts are present, but also how multiple concepts jointly manifest in the visual content. We therefore hypothesize that intermediate diffusion representations encode semantic and structural cues complementary to CLIP's contrastive embeddings, providing a useful prior for modeling attribute--object interactions and relational structure across compositions. Guided by this hypothesis, we investigate diffusion representations as supplementary semantic priors for enhancing CLIP-based compositional representations.

Motivated by this perspective, as illustrated in Fig.~\ref{fig:intro}(d), we propose \textbf{DIFFCZSL}, a diffusion-augmented framework that regularizes CLIP-based CZSL models with diffusion-derived semantic priors. Rather than replacing the discriminative backbone, DIFFCZSL introduces diffusion representations as auxiliary training signals to reshape the compositional geometry of both image and text embeddings.
In this way, the model can benefit from complementary compositional cues encoded by generative representations while preserving the efficient CLIP-based inference pipeline. Concretely, we extract intermediate features from a pre-trained diffusion model and project them into the CLIP embedding space via lightweight heads. The projected diffusion representations then provide auxiliary alignment targets for both image embeddings and compositional text embeddings during training. Moreover, DIFFCZSL is backbone-agnostic and can be readily integrated into representative CLIP-based CZSL pipelines without altering their core inference architectures.

In summary, our contributions are threefold: \textbf{(i)} We identify two limitations of contrastively pre-trained representations for CZSL: limited modeling of attribute--object interactions within individual compositions and limited relational consistency across compositions, both of which can hinder generalization to unseen attribute--object combinations. \textbf{(ii)} We propose \textbf{DIFFCZSL}, a diffusion-augmented framework that leverages composition-aware diffusion representations as auxiliary priors to regularize both visual and textual representations during training, while introducing no additional diffusion-model cost at inference time. \textbf{(iii)} We conduct extensive experiments on three standard CZSL benchmarks, MIT-States~\citep{mit}, UT-Zappos~\citep{ut}, and C-GQA~\citep{cgqa}, demonstrating consistent improvements across representative CLIP-based CZSL models under both closed-world and open-world settings.

\section{Related Work}
\subsection{Compositional Zero-Shot Learning (CZSL)}
Compositional Zero-Shot learning (CZSL) focuses on generalizing to novel attribute–object compositions by learning transferable representations of primitive concepts from seen compositions. Early CZSL methods can be broadly categorized into two paradigms. The first learns holistic representations for attribute–object compositions and maps images and compositions into a shared semantic space, so that unseen compositions can be recognized through compositional synthesis~\citep{misra2017red,cgqa,mancini2021open,mancini2022learning}. The second explicitly disentangles visual representations into state and object components, models the two primitives with separate branches or classifiers, and then combines their predictions for final composition recognition~\citep{li2022siamese,li2020symmetry,kim2023hierarchical}.

More recently, large-scale vision--language models (VLMs) such as CLIP~\citep{radford2021learning} have significantly advanced CZSL by enabling zero-shot recognition through cross-modal alignment. Existing CLIP-based methods typically adopt parameter-efficient fine-tuning strategies to improve compositional generalization~\citep{csp,He2026FlowComposer,huang2024troika,cams}. Specifically, they introduce learnable compositional prompts, branch-specific representations, or lightweight visual adapters to model attributes, objects, and their interactions while retaining most of CLIP's pretrained parameters. Nevertheless, these methods remain largely dependent on the contrastive representations learned from image and text pairs, which may not fully capture the compositional relationships between attributes and objects. This limitation motivates the exploration of complementary supervision with richer compositional semantics. In this work, we investigate diffusion representations as such a complementary signal for enhancing compositional generalization in CZSL.
 
\subsection{Diffusion Representations for Visual Recognition}

Diffusion models were originally developed for image generation, evolving from denoising diffusion probabilistic models (DDPMs)~\citep{ho2020denoising,dhariwal2021diffusion} to large-scale text-to-image systems such as Stable Diffusion~\citep{rombach2022high}. Beyond their generative capabilities, recent studies have shown that intermediate diffusion features can serve as effective visual representations. These features encode semantic and structural information that transfers to a broad range of recognition tasks, including zero-shot classification~\citep{clark2023text,li2023your,yang2023diffusion}, semantic segmentation~\citep{baranchuk2021label,xu2023open}, and semantic correspondence~\citep{he2023discffusion}. CleanDIFT ~\citep{stracke2025cleandift} further demonstrates that stable and semantically meaningful representations can be extracted directly from clean images without relying on noisy denoising trajectories.

Beyond general visual recognition, emerging evidence suggests that diffusion representations are also effective for compositional reasoning. For example, \citet{li2023your} show that diffusion models can function as zero-shot classifiers and outperform contrastively trained vision--language models on multimodal compositional reasoning tasks. Unlike contrastive learning, which primarily optimizes global image--text matching, text-conditioned denoising requires the model to integrate multiple textual concepts with the corresponding visual content. This training objective may encourage intermediate diffusion features to preserve joint semantic relationships that are useful for recognizing attribute--object compositions.

Several works have explored representation transfer between diffusion and discriminative models. RepFusion~\citep{yang2023diffusion} distills diffusion representations through dynamic timestep optimization, while DreamTeacher~\citep{li2023dreamteacher} transfers diffusion knowledge to downstream visual backbones through cross-model feature distillation. DDAE~\citep{xiang2023denoising} reports a connection between generative quality and representation learning performance, further supporting the discriminative value of diffusion-derived features. In the reverse direction, REPA~\citep{yu2025repa} shows through linear probing that diffusion-transformer hidden states contain meaningful discriminative information, although their quality remains below that of DINOv2 features. It therefore uses clean-image representations from DINOv2 as semantic targets to align noisy diffusion hidden states, accelerating diffusion training and improving generation quality. However, these studies mainly focus on general-purpose representation transfer or diffusion training efficiency and do not investigate how diffusion features can support compositional generalization in CZSL.

Different from prior efforts that mainly focus on efficiency or replacing the original model, our work leverages diffusion representations as compositional priors for CZSL. Rather than replacing the backbone, we inject diffusion features into CLIP-based CZSL pipelines as auxiliary supervision during training, with the goal of reshaping the embedding geometry toward more composition-aware semantics.

\section{Preliminaries}
\subsection{Problem Statement} 
CZSL seeks to generalize to unseen attribute–object compositions by leveraging primitive concepts learned from seen compositions. Given a state set $\mathcal{A}=\{a_1,a_2,\dots,a_{|\mathcal{A}|}\}$ and an object set $\mathcal{O}=\{o_1,o_2,\dots,o_{|\mathcal{O}|}\}$, the compositional label space is defined as $\mathcal{C}=\mathcal{A}\times\mathcal{O}$, where each composition \( c = (a, o) \) corresponds to a state--object pair. The composition space is further partitioned into two disjoint subsets: the seen set \( \mathcal{C}^{se} \) and the unseen set \( \mathcal{C}^{us} \), such that $\mathcal{C}^{se} \cap \mathcal{C}^{us} = \varnothing$. During training, a dataset $\mathcal{D}_T = \{(x_i, c_i) \mid x_i \in \mathcal{X},\, c_i \in \mathcal{C}^{se}\}$ is provided, where \( \mathcal{X} \) denotes the image space. At inference time, the model is required to assign compositional labels from a target set \( \mathcal{C}^{tgt} \). In the \emph{closed-world} setting, the target set is restricted to $\mathcal{C}^{tgt} = \mathcal{C}^{se} \cup \mathcal{C}^{us}$, where only the known composition space is considered. In the \emph{open-world} setting, the target set includes all possible state--object compositions, $\mathcal{C}^{tgt} = \mathcal{C}$.

\subsection{Baselines Review}

We categorize existing CLIP-based CZSL methods into two groups: \emph{single-path} and \emph{multi-path} formulations. Both are built upon contrastively trained vision--language models,  but differ in how compositional representations are constructed.

\noindent \textbf{Single-path Baseline.}
CSP~\citep{csp} formulates CZSL as a single-path alignment between image embeddings and compositional text embeddings. Both CLIP encoders are kept frozen, and only a set of learnable soft tokens is optimized. These tokens represent attribute and object concepts and are inserted into a fixed prompt template (\eg, ``a photo of [attribute] [object]''). For each attribute--object pair, the model computes temperature-scaled cosine similarity between the image feature and the corresponding compositional prompt embedding, followed by a softmax over all candidate compositions. Training is performed using a standard cross-entropy objective.

\noindent \textbf{Multi-path Baseline.}
Multi-path methods, such as Troika~\citep{huang2024troika} and CAMS~\citep{cams}, explicitly decompose compositional reasoning into multiple branches. In particular, they adopt a three-path architecture that models attribute, object, and composition representations separately. On the visual side, lightweight adapters and disentanglement modules are introduced to extract branch-specific features from the shared image representation. On the text side, learnable primitive tokens are shared across branches, together with branch-specific prompt prefixes for different concepts. Each branch produces predictions based on cosine similarity between image and text embeddings, and the model is trained jointly across all branches using cross-entropy supervision. During inference, the final composition prediction is obtained by combining compositional and primitive-level scores.

\begin{figure}[!t]
    \centering
    \includegraphics[
        width=\linewidth,
        height=0.75\textheight,
        keepaspectratio
    ]{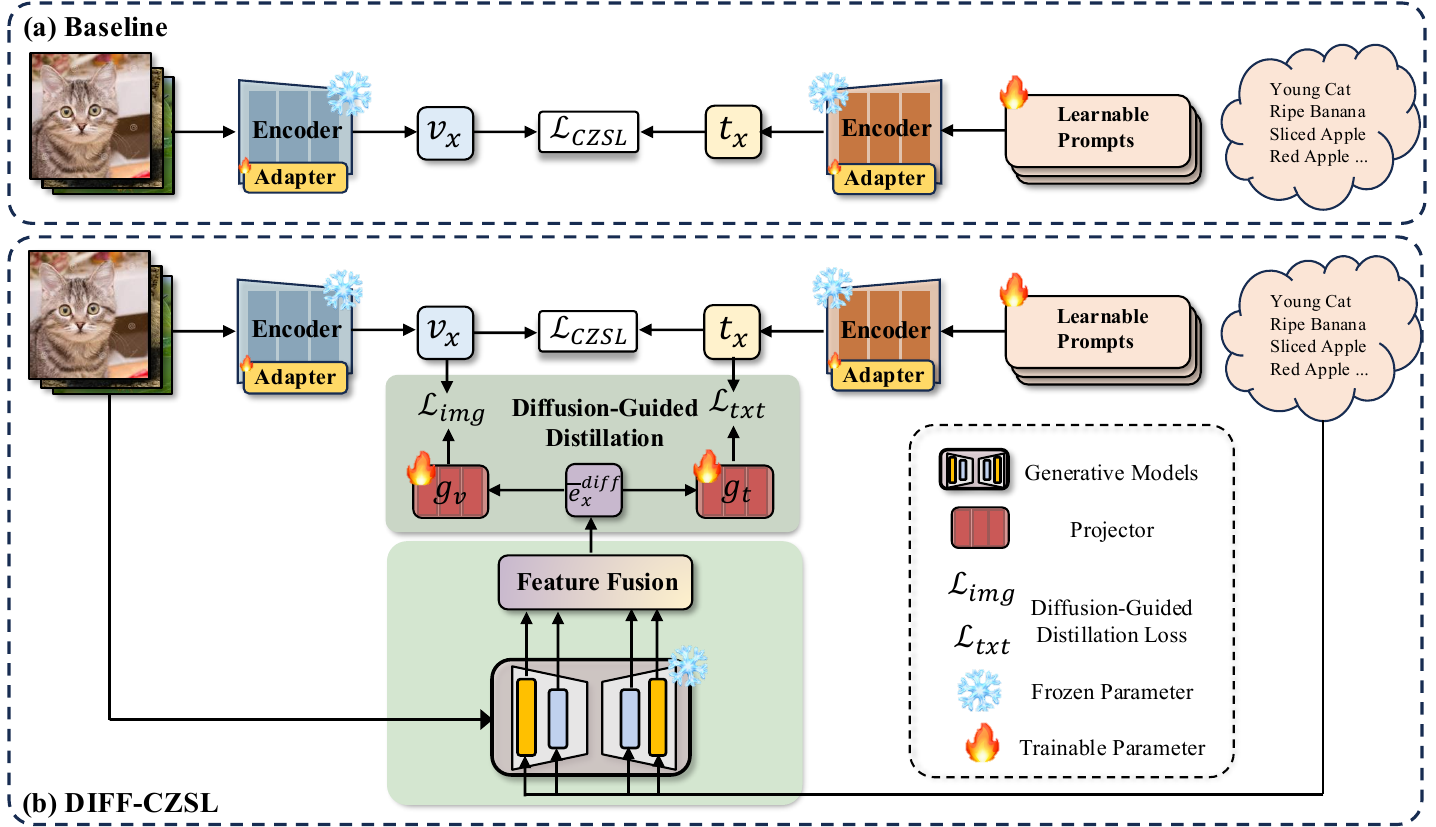}
    \caption{
    Overview of our DIFFCZSL framework.
    (a) Existing CLIP-based CZSL framework.
    (b) DIFF-CZSL introduces diffusion-derived features as auxiliary supervision during training.
    The diffusion branch is removed at inference time, incurring no extra inference cost.
    }
    \label{fig:framework}
\end{figure}

\section{Method: DIFFCZSL}
\noindent\textbf{Overview.}
CLIP-based CZSL models perform recognition in a contrastively pretrained image--text embedding space, which may not fully capture  relationships between attributes and objects. To complement these representations, we propose \textbf{DIFFCZSL}, a plug-and-play framework that uses intermediate diffusion features as auxiliary compositional priors during training. As illustrated in Fig.~\ref{fig:framework}, DIFFCZSL first extracts intermediate features from a frozen diffusion model, projects them into the visual and textual embedding spaces of the target CZSL model, and then aligns them with the corresponding image and compositional text representations through two distillation objectives. This joint regularization transfers diffusion-derived semantic information to both sides of the shared embedding space. Since the diffusion branch is removed after training, the original architecture and inference procedure of the baseline remain unchanged.

\subsection{Diffusion Features as Compositional Priors}

During training, given a labeled image--composition pair $(x,(a,o))$, we construct an attribute--object text condition $t_{a,o}$ using the corresponding attribute and object names. We feed both $x$ and $t_{a,o}$ into the frozen Stable Diffusion feature extractor, whose cross-attention layers integrate the compositional semantics of $t_{a,o}$ with the visual information of $x$~\citep{rombach2022high}. The resulting intermediate representation is a text-conditioned spatial feature tensor:
\begin{equation}
e_x^{\mathrm{diff}} \in \mathbb{R}^{C \times H \times W},
\end{equation}
where $C$ denotes the channel dimension and $H \times W$ is the spatial resolution. Since compositional information is distributed across local regions, we aggregate the spatial feature into a compact global representation by average pooling:
\begin{equation}
\bar{e}_x^{\mathrm{diff}}
=
\frac{1}{HW}
\sum_{h=1}^{H}\sum_{w=1}^{W}
e_x^{\mathrm{diff}}(h,w).
\end{equation}

However, directly using diffusion features for CZSL is suboptimal due to the inherent representation gap between generative diffusion models and discriminative CZSL backbones. The two models are trained with fundamentally different objectives and therefore organize semantic information in different embedding geometries. To make diffusion priors usable for CZSL, we further introduce a learnable projection mechanism that adapts them into the target embedding spaces of the baseline model.

Specifically, we employ two projection heads, $g_v(\cdot)$ and $g_t(\cdot)$, for both image and text sides, respectively, to map the pooled diffusion prior into the visual and textual spaces of the CZSL model:
\begin{equation}
\mathbf{v}_x^{\mathrm{diff}} = g_v(\bar{e}_x^{\mathrm{diff}}) ,\quad
\mathbf{t}_x^{\mathrm{diff}} = g_t(\bar{e}_x^{\mathrm{diff}}).
\end{equation}
Here, $\mathbf{v}_x^{\mathrm{diff}}$ provides a diffusion-guided compositional target for the image representation, while $\mathbf{t}_x^{\mathrm{diff}}$ serves as a corresponding supervisory signal in the compositional text space. This design allows the baseline to absorb diffusion-induced relational structure from both visual and semantic perspectives, rather than merely matching isolated features in a single space.

In this way, our method does not simply use diffusion as an auxiliary encoder; instead, it leverages diffusion features as transferable compositional priors and injects them into the training of discriminative CZSL models. Importantly, this transfer happens only during training, so the original architecture and inference procedure of the baseline remain unchanged.

\subsection{Compositional Distillation}

With the projected diffusion priors in hand, the next step is to transfer their compositional structure into the target CZSL model. Rather than directly replacing the original visual or textual representations, we use the diffusion priors as an auxiliary teacher to regularize the learning of the baseline. In this way, the CZSL model is encouraged to organize its embedding space not only according to discriminative supervision, but also according to the compositional relations encoded by the pre-trained diffusion model.

Let $\mathbf{v}_x$ denote the image embedding produced by the CZSL backbone for image $x$, and let $\mathbf{t}_{a,o}$ denote the text embedding of the corresponding attribute--object composition $(a,o)$. From the diffusion prior branch, we obtain the projected visual prior $\mathbf{v}_x^{\mathrm{diff}}$ and textual prior $\mathbf{t}_x^{\mathrm{diff}}$ as defined in the previous subsection. We then distill the compositional knowledge carried by these priors into the baseline model through two complementary objectives.

\noindent\textbf{Dual-Space Compositional Distillation.}
We transfer diffusion-derived compositional information into both the visual and textual representation spaces of the CZSL model. Let $\mathcal{M}=\{\mathrm{img},\mathrm{txt}\}$ denote the two alignment spaces. For each sample $i$ in a mini-batch of size $B$, we define the diffusion prior $\mathbf{d}_i^{(m)}$ and its corresponding baseline representation $\mathbf{z}_i^{(m)}$ as
\begin{equation}
\left(\mathbf{d}_i^{(m)},\mathbf{z}_i^{(m)}\right)
=
\begin{cases}
\left(\mathbf{v}_i^{\mathrm{diff}},\mathbf{v}_i\right),
& m=\mathrm{img}, \\[2pt]
\left(\mathbf{t}_i^{\mathrm{diff}},\mathbf{t}_{i}\right),
& m=\mathrm{txt},
\end{cases}
\end{equation}

where $\mathbf{v}_i$ is the baseline image embedding, $\mathbf{t}_{i}$ is the text embedding of its corresponding
attribute--object composition, and $\mathbf{v}_i^{\mathrm{diff}}$ and $\mathbf{t}_i^{\mathrm{diff}}$ are the diffusion priors projected into the respective embedding spaces.

For each space $m\in\mathcal{M}$, we compute the cross-view similarity matrix as
\begin{equation}
S_{ij}^{(m)}
=
\frac{
\left(\mathbf{d}_i^{(m)}\right)^{\top}
\mathbf{z}_j^{(m)}
}{\tau},
\end{equation}
where $\tau$ is the temperature hyperparameter. Each diffusion prior is paired with the corresponding baseline representation, making the diagonal entries positive pairs and the off-diagonal entries negative pairs. We then apply the same bidirectional contrastive objective in both spaces:

\begin{equation}
\mathcal{L}_{m}
=
-\frac{1}{2B}
\sum_{i=1}^{B}
\left[
\log
\frac{\exp\left(S_{ii}^{(m)}\right)}
{\sum_{j=1}^{B}\exp\left(S_{ij}^{(m)}\right)}
+
\log
\frac{\exp\left(S_{ii}^{(m)}\right)}
{\sum_{j=1}^{B}\exp\left(S_{ji}^{(m)}\right)}
\right],
\qquad
m\in\mathcal{M}.
\end{equation}

Specifically, $\mathcal{L}_{\mathrm{img}}$ and $\mathcal{L}_{\mathrm{txt}}$ regularize the image and compositional text embedding spaces, respectively, allowing diffusion priors to constrain both sides of the shared cross-modal space.

\noindent\textbf{Overall training objective.}
Finally, we combine the original training loss of the CZSL baseline, denoted as $\mathcal{L}_{\mathrm{CZSL}}$, with the two diffusion-guided distillation terms:
\begin{equation}
\mathcal{L}
=
\mathcal{L}_{\mathrm{CZSL}}
+
\lambda_{\mathrm{img}} \mathcal{L}_{\mathrm{img}}
+
\lambda_{\mathrm{txt}} \mathcal{L}_{\mathrm{txt}},
\end{equation}
where $\lambda_{\mathrm{img}}$ and $\lambda_{\mathrm{txt}}$ control the strengths of visual and textual distillation, respectively.

Overall, our approach augments the CZSL baseline with diffusion-derived compositional priors by distilling them into both visual and semantic branches during training. Importantly, the diffusion branch is used only for auxiliary supervision, leaving the original architecture and inference pipeline unchanged.

\section{Experiments}
\subsection{Experiment Settings}
\noindent \textbf{Benchmark.}  We evaluate our method on three standard CZSL benchmarks. \textit{MIT-States}~\citep{mit} contains $53,753$ images with $115$ attributes and $245$ object categories. \textit{UT-Zappos50K}~\citep{ut} contains $50,025$ images with $16$ attributes and $12$ shoe categories. \textit{C-GQA}~\citep{cgqa} contains $39,298$ images with a diverse collection of attribute and object categories covering common real-world concepts.
Following the standard CZSL evaluation setup~\citep{csp,huang2024troika}, we introduce a calibration bias to the scores of \emph{unseen} compositions and sweep it over the range from $-\infty$ to $+\infty$ to balance seen and unseen predictions. We then report the Area Under the seen--unseen trade-off Curve (\textbf{AUC}) and the best harmonic mean (\textbf{HM}) over the sweep. We also include the maximum seen accuracy (\textbf{Seen}) and maximum unseen accuracy (\textbf{Unseen}) achieved during this process. For open-world evaluation, we follow~\cite{csp} and apply a post-training calibration step to remove infeasible compositions.

\noindent\textbf{Baselines.}
We compare our method with a broad range of approaches. We first include CLIP~\citep{radford2021learning} and CoOp~\citep{zhou2022learning} as general vision--language baselines. We then compare with representative CZSL methods, including CSP~\citep{csp}, DFSP~\citep{lu2023decomposed}, DLM~\citep{hu2024dynamic}, Troika~\citep{huang2024troika}, CDS-CZSL~\citep{li2024context}, IMAX~\citep{jiang2024imaginary}, and CAMS~\citep{cams}. These methods cover several major directions in CZSL, including compositional prompt learning, cross-modal feature fusion, primitive-level modeling, and composition-aware representation learning. We additionally include PLID~\citep{PLID}, PLO~\citep{li2025compositional}, and LOGICZSL~\citep{wu2025logiczsl}, which leverage external knowledge from large language models to enhance compositional reasoning.

\noindent\textbf{Implementation Details.}
We implement our framework in PyTorch and follow the training setup of prior work~\citep{huang2024troika}. DIFFCZSL is designed as an auxiliary representation-level regularizer that can be plugged into existing CLIP-based CZSL models during training without modifying their core architectures or inference procedures. To demonstrate its plug-and-play compatibility and effectiveness across different model designs, we integrate DIFFCZSL into three representative CZSL approaches: CSP~\citep{csp}, Troika~\citep{huang2024troika}, and CAMS~\citep{cams}, resulting in DIFF-CSP, DIFF-Troika, and DIFF-CAMS, respectively. CSP represents a single-path compositional prompt-learning framework, while Troika adopts a multi-path architecture that separately models attributes, objects, and their compositions. CAMS is a recent high-performing CZSL method that further enhances multi-path compositional modeling. We use the pretrained CLIP ViT-L/14~\citep{radford2021learning} as both the image and text encoder and adopt CleanDIFT~\citep{stracke2025cleandift}, built upon Stable Diffusion 2.1, for diffusion-based feature extraction. Since the diffusion branch is used only during training, DIFFCZSL preserves the original inference pipeline and introduces no additional inference-time computational cost. Detailed implementation details are listed in Appendix~\ref{app:implementation}.

\begin{table*}[t]

  \caption{Quantitative comparison (\S~\ref{sec:main_results}) on three benchmarks under closed-world and open-world settings. * denotes results from our implementation. $\ddagger$ represents the methods leveraging the LLM's knowledge.}
  \label{tab:main_results}
  \centering
  \setlength{\tabcolsep}{2.0pt}{
  \resizebox{1\linewidth}{!}{
\begin{tabular}{l|cccc|cccc|cccc}
\toprule
\hline
&\multicolumn{4}{c|}{MIT-States~\citep{mit}}& \multicolumn{4}{c|}{UT-Zappos~\citep{ut}}&\multicolumn{4}{c}{C-GQA~\citep{cgqa}}\\
Method&Seen&Unseen&HM&AUC&Seen&Unseen&HM&AUC&Seen&Unseen&HM&AUC\\
\hline
\multicolumn{13}{c}{\textbf{\textit{Closed-world Results}}}\\
\hline 

CLIP \scriptsize\textcolor{gray}{[ICML21]}& 30.2  & 46.0  & 26.1  & 11.0  & 15.8  & 49.1  & 15.6  & 5.0   & 7.5   & 25.0  & 8.6   & 1.4 \\
CoOP \scriptsize\textcolor{gray}{[IJCV22]} & 34.4  & 47.6  & 29.8  & 13.5  & 52.1  & 49.3  & 34.6  & 18.8  & 20.5  & 26.8  & 17.1  & 4.4 \\
DFSP(i2t) \scriptsize\textcolor{gray}{[CVPR23]} & 47.4  & 52.4  & 37.2  & 20.7  & 64.2  & 66.4  & 45.1  & 32.1  & 35.6  & 29.3  & 24.3  & 8.7 \\
DFSP(BiF) \scriptsize\textcolor{gray}{[CVPR23]} & 47.1  & 52.8  & 37.7  & 20.8  & 63.3  & 69.2  & 47.1  & 33.5  & 36.5  & 32.0  & 26.2  & 9.9 \\
DFSP(t2i) \scriptsize \textcolor{gray}{[CVPR23]} & 46.9  & 52.0  & 37.3  & 20.6  & 66.7  & 71.7  & 47.2  & 36.0  & 38.2  & 32.0  & 27.1  & 10.5 \\
DLM \scriptsize\textcolor{gray}{[AAAI24]}   & 46.3 & 49.8 & 37.4 & 20.0   & 67.1 & 72.5 & 52.0 & 39.6  & 32.4 & 28.5 & 21.9 & 7.3  \\ 
CDS-CZSL  \scriptsize \textcolor{gray}{[CVPR24]}   &50.3 & 52.9 & 39.2 & 22.4    & 63.9 & 74.8 & 52.7 & 39.5    & 38.3 & 34.2 & 28.1 & 11.1     \\ 
IMAX                  \scriptsize \textcolor{gray}{[TPAMI25]}& 48.7& 53.8  & 39.1 & 21.9& 69.3&70.7 & 54.2 & 40.6   &39.7 & 35.8& 29.8 & 12.8      \\ 
PLID$^\ddagger$   \scriptsize \textcolor{gray}{[ECCV24]} & 49.7 & 52.4 & 39.0 & 22.1   & 67.3 & 68.8 & 52.4 & 38.7 & 38.8 & 33.0 & 27.9 & 11.0    \\
PLO$^\ddagger$ \scriptsize \textcolor{gray}{[ACMMM25]} & 51.6 & 53.7 & 40.2 & 23.4 & 70.3 & 75.8 & 55.3 & 43.6 &  44.7 & 38.1 & 33.0 & 14.9\\
LOGICZSL$^\ddagger$ \scriptsize \textcolor{gray}{[CVPR25]} & 50.8& 53.9& 40.5& 23.4 &69.6& 74.9 &57.8& 45.8& 44.4 &39.4 &33.3 &15.3 \\
\hline
CSP$^*$\scriptsize \textcolor{gray}{[ICLR23]} & 47.1& 49.0 & 36.1&19.3 & 64.2&65.5&46.6&33.0&28.7&26.8&19.7&5.8\\

\cellcolor{my_blue}\textbf{DIFF-CSP}&\cellcolor{my_blue}47.3 $\textcolor{blue}{\uparrow\scriptsize 0.2}$& \cellcolor{my_blue}49.5$\textcolor{blue}{\uparrow\scriptsize 0.5}$  & \cellcolor{my_blue}36.6 $\textcolor{blue}{\uparrow\scriptsize 0.5}$&\cellcolor{my_blue}19.6 $\textcolor{blue}{\uparrow\scriptsize 0.3}$ &\cellcolor{my_blue}65.7 $\textcolor{blue}{\uparrow\scriptsize 1.5}$&\cellcolor{my_blue}66.4$\textcolor{blue}{\uparrow\scriptsize 0.9}$&\cellcolor{my_blue}47.6 $\textcolor{blue}{\uparrow\scriptsize 1.0}$&\cellcolor{my_blue}34.4 $\textcolor{blue}{\uparrow\scriptsize 1.4}$&\cellcolor{my_blue}29.5$\textcolor{blue}{\uparrow\scriptsize 0.8}$&\cellcolor{my_blue}27.1$\textcolor{blue}{\uparrow\scriptsize 0.3}$&\cellcolor{my_blue}20.5$\textcolor{blue}{\uparrow\scriptsize 0.8}$\cellcolor{my_blue}&\cellcolor{my_blue}6.4$\textcolor{blue}{\uparrow\scriptsize 0.6}$\\
\hline
Troika$^*$ \scriptsize \textcolor{gray}{[CVPR24]}&49.0 &53.0 &39.2&22.1& 66.8&73.4&54.6&41.7&41.0&35.7&29.4&12.4 \\

\cellcolor{my_blue}\textbf{DIFF-Troika}& \cellcolor{my_blue}50.5$\textcolor{blue}{\uparrow\scriptsize 1.5}$
& \cellcolor{my_blue}53.4 $\textcolor{blue}{\uparrow\scriptsize 0.4}$& \cellcolor{my_blue}39.9$\textcolor{blue}{\uparrow\scriptsize 0.7}$&\cellcolor{my_blue}23.1$\textcolor{blue}{\uparrow\scriptsize 1.0}$ & \cellcolor{my_blue}72.2$\textcolor{blue}{\uparrow\scriptsize 5.4 }$&\cellcolor{my_blue}74.0$\textcolor{blue}{\uparrow\scriptsize 0.6}$&\cellcolor{my_blue}57.6$\textcolor{blue}{\uparrow\scriptsize 3.0}$&\cellcolor{my_blue}46.2$\textcolor{blue}{\uparrow\scriptsize 4.5}$&\cellcolor{my_blue}43.0$\textcolor{blue}{\uparrow\scriptsize 2.0}$&\cellcolor{my_blue}38.8$\textcolor{blue}{\uparrow\scriptsize 3.1}$&\cellcolor{my_blue}32.0$\textcolor{blue}{\uparrow\scriptsize 2.6}$&\cellcolor{my_blue}14.3$\textcolor{blue}{\uparrow\scriptsize 1.9}$\\
\hline

CAMS$^*$ & 52.8&53.2&40.3&23.4& 69.4&76.2&56.6&45.0&45.6&39.8&33.7&15.8 \\

\cellcolor{my_blue}\textbf{DIFF-CAMS}& \cellcolor{my_blue}52.5$\textcolor{black}{\downarrow\scriptsize 0.3}$&\cellcolor{my_blue}53.9$\textcolor{blue}{\uparrow\scriptsize 0.7}$&\cellcolor{my_blue}40.7$\textcolor{blue}{\uparrow\scriptsize 0.4}$&\cellcolor{my_blue}24.0$\textcolor{blue}{\uparrow\scriptsize 0.6}$& \cellcolor{my_blue}70.3$\textcolor{blue}{\uparrow\scriptsize 0.9}$&\cellcolor{my_blue}76.4 $\textcolor{blue}{\uparrow\scriptsize 0.2}$ &\cellcolor{my_blue}57.9$\textcolor{blue}{\uparrow\scriptsize 1.3}$&\cellcolor{my_blue}46.1$\textcolor{blue}{\uparrow\scriptsize 1.1}$&\cellcolor{my_blue}46.0$\textcolor{blue}{\uparrow\scriptsize 0.4}$&\cellcolor{my_blue}39.8$\textcolor{gray}{\rightarrow\scriptsize 0.0}$&\cellcolor{my_blue}34.2$\textcolor{blue}{\uparrow\scriptsize 0.5}$&\cellcolor{my_blue}16.2$\textcolor{blue}{\uparrow\scriptsize 0.4}$ \\

\hline
\multicolumn{13}{c}{\textbf{\textit{Open-world Results}}}\\
\hline
CLIP\scriptsize \textcolor{gray}{[ICML21]}& 30.1  & 14.3  & 12.8  & 3.0   & 15.7  & 20.6  & 11.2  & 2.2   & 7.5   & 4.6   & 4.0   & 0.3 \\
CoOP\scriptsize \textcolor{gray}{[IJCV22]} & 34.6  & 9.3   & 12.3  & 2.8   & 52.1  & 31.5  & 28.9  & 13.2  & 21.0  & 4.6   & 5.5   & 0.7 \\
DFSP(i2t)\scriptsize \textcolor{gray}{[CVPR23]}& 47.2  & 18.2  & 19.1  & 6.7   & 64.3  & 53.8  & 41.2  & 26.4  & 35.6  & 6.5   & 9.0   & 2.0 \\
DFSP(BiF)\scriptsize \textcolor{gray}{[CVPR23]} & 47.1  & 18.1  & 19.2  & 6.7   & 63.5  & 57.2  & 42.7  & 27.6  & 36.4  & 7.6   & 10.6  & 2.4 \\
DFSP(t2i)\scriptsize \textcolor{gray}{[CVPR23]} & 47.5  & 18.5  & 19.3  & 6.8   & 66.8 & 60.0  & 44.0  & 30.3  & 38.3  & 7.2   & 10.4  & 2.4  \\
CDS-CZSL \scriptsize \textcolor{gray}{[CVPR24]} & 49.4 & 21.8 & 22.1 & 8.5 & 64.7 & 61.3 & 48.2 & 32.3 & 37.6 & 8.2 & 11.6 & 2.7   \\
IMAX \scriptsize \textcolor{gray}{[TPAMI25]} & 50.2& 18.6  & 21.4 & 7.6 & 68.4&57.3 & 47.5 & 32.3   &38.7 & 7.9& 11.2 & 2.5    \\ 
PLID$^\ddagger$   \scriptsize\textcolor{gray}{[ECCV24]}& 49.1 & 18.7 & 20.4 & 7.3& 67.6 & 55.5 & 46.6 & 30.8 &39.1 & 7.5 & 10.6 & 2.5 \\   
PLO$^\ddagger$\scriptsize\textcolor{gray}{[ACMMM25]} & 49.7 & 19.4 & 21.4 & 7.8 & 68.0 & 63.5 & 47.8 & 33.1 & 43.9 & 10.4 & 13.9 & 3.9 \\
LOGICZSL$^\ddagger$ \scriptsize \textcolor{gray}{[CVPR25]} & 50.7 &21.4& 22.4& 8.7 &69.6& 63.7 &50.8 &36.2& 43.7& 9.3& 12.6 &3.4\\

\hline
CSP$^*$\scriptsize\textcolor{gray}{[ICLR23]} &  47.1&15.3&17.0&5.6
 &  64.1&44.1&38.9&22.7&28.7&5.2&7.0&1.2\\
\cellcolor{my_blue}\textbf{DIFF-CSP}& \cellcolor{my_blue}47.3$\textcolor{blue}{\uparrow\scriptsize 0.2}$&\cellcolor{my_blue}15.5$\textcolor{blue}{\uparrow\scriptsize 0.2}$&\cellcolor{my_blue}17.1$\textcolor{blue}{\uparrow\scriptsize 0.1}$\cellcolor{my_blue}&\cellcolor{my_blue}5.7 $\textcolor{blue}{\uparrow\scriptsize 0.1}$& \cellcolor{my_blue}65.6$\textcolor{blue}{\uparrow\scriptsize 1.5}$&\cellcolor{my_blue}45.0$\textcolor{blue}{\uparrow\scriptsize 0.9}$&\cellcolor{my_blue}39.4$\textcolor{blue}{\uparrow\scriptsize 0.5}$&\cellcolor{my_blue}23.5$\textcolor{blue}{\uparrow\scriptsize 0.8}$&\cellcolor{my_blue}29.4$\textcolor{blue}{\uparrow\scriptsize 0.7}$&\cellcolor{my_blue}5.4$\textcolor{blue}{\uparrow\scriptsize 0.2}$&\cellcolor{my_blue}7.1$\textcolor{blue}{\uparrow\scriptsize 0.1}$&\cellcolor{my_blue}1.4$\textcolor{blue}{\uparrow\scriptsize 0.2}$ \\

\hline
Troika$^*$\scriptsize \textcolor{gray}{[CVPR24]}& 48.8&17.5& 19.1&6.8&66.4&61.0&47.8&33.0& 40.8  &7.9 &10.9 &2.7 \\
\cellcolor{my_blue}\textbf{DIFF-Troika}& \cellcolor{my_blue}50.5 $\textcolor{blue}{\uparrow\scriptsize 1.7}$&\cellcolor{my_blue}19.6$\textcolor{blue}{\uparrow\scriptsize 2.1}$&\cellcolor{my_blue}20.7$\textcolor{blue}{\uparrow\scriptsize 1.6}$&\cellcolor{my_blue}7.8$\textcolor{blue}{\uparrow\scriptsize 1.0}$& \cellcolor{my_blue}72.2$\textcolor{blue}{\uparrow\scriptsize 5.8}$&\cellcolor{my_blue}61.4$\textcolor{blue}{\uparrow\scriptsize 0.4}$&\cellcolor{my_blue}50.1$\textcolor{blue}{\uparrow\scriptsize 2.3}$ &\cellcolor{my_blue}36.6$\textcolor{blue}{\uparrow\scriptsize 3.6}$&\cellcolor{my_blue}43.0$\textcolor{blue}{\uparrow\scriptsize 2.2}$&\cellcolor{my_blue}8.5$\textcolor{blue}{\uparrow\scriptsize 0.6}$&\cellcolor{my_blue}11.6$\textcolor{blue}{\uparrow\scriptsize 0.7}$&\cellcolor{my_blue}3.1 $\textcolor{blue}{\uparrow\scriptsize 0.4}$ \\
\hline

CAMS$^{*}$ & 52.7&20.9&22.0&8.5& 69.4&61.6&50.8&35.7&45.7&10.6&14.4&4.1 \\
\cellcolor{my_blue}\textbf{DIFF-CAMS}& \cellcolor{my_blue}52.5$\textcolor{black}{\downarrow\scriptsize 0.2}$&\cellcolor{my_blue}21.1$\textcolor{blue}{\uparrow\scriptsize 0.2}$&\cellcolor{my_blue}22.1$\textcolor{blue}{\uparrow\scriptsize 0.1}$&\cellcolor{my_blue}8.7$\textcolor{blue}{\uparrow\scriptsize 0.2}$& \cellcolor{my_blue}70.1$\textcolor{blue}{\uparrow\scriptsize 0.7}$&\cellcolor{my_blue}64.1$\textcolor{blue}{\uparrow\scriptsize 2.5}$&\cellcolor{my_blue}52.6$\textcolor{blue}{\uparrow\scriptsize 1.8}$&\cellcolor{my_blue}38.1$\textcolor{blue}{\uparrow\scriptsize 2.4}$&\cellcolor{my_blue}46.0$\textcolor{blue}{\uparrow\scriptsize 0.3}$&\cellcolor{my_blue}10.9$\textcolor{blue}{\uparrow\scriptsize 0.3}$&\cellcolor{my_blue}14.5$\textcolor{blue}{\uparrow\scriptsize 0.1}$&\cellcolor{my_blue}4.3$\textcolor{blue}{\uparrow\scriptsize 0.2}$ \\
\bottomrule
\end{tabular}
}
}
\end{table*}

\subsection{Main Results}
\label{sec:main_results}
Table~\ref{tab:main_results} reports quantitative comparisons with recent methods on three datasets: MIT-States~\citep{mit}, UT-Zappos~\citep{ut}, and C-GQA~\citep{cgqa}, under both closed-world and open-world settings. Integrating our DIFFCZSL framework into CSP~\citep{csp}, Troika~\citep{huang2024troika}, and CAMS~\citep{cams} yields consistent improvements across most of the evaluation metrics on all datasets, demonstrating the robustness and model-agnostic nature of our formulation.

\noindent\textbf{MIT-States.}
As shown in Tab.~\ref{tab:main_results} (columns 2--5), all three diffusion-augmented models improve HM and AUC over their corresponding baselines on MIT-States under both closed-world and open-world settings. The most notable gains are obtained with Troika, where DIFF-Troika improves HM and AUC from 39.2 and 22.1 to 39.9 and 23.1 in the closed-world setting, and from 19.1 and 6.8 to 20.7 and 7.8 in the open-world setting. DIFF-CAMS exhibits a slightly different behavior: its maximum Seen accuracy decreases from 52.8 to 52.5 in the closed-world setting and from 52.7 to 52.5 in the open-world setting. However, these small decreases are accompanied by improvements in Unseen accuracy, HM, and AUC. In particular, its closed-world HM and AUC increase from 40.3 and 23.4 to 40.7 and 24.0. The higher AUC indicates an improved overall trade-off between seen and unseen recognition across different calibration biases, while the increased HM reflects a better balance at the optimal operating point. Therefore, despite a slight decrease in maximum Seen accuracy, DIFF-CAMS achieves higher Unseen accuracy and a better overall seen and unseen trade-off.

\noindent\textbf{UT-Zappos.}
Tab.~\ref{tab:main_results} (columns 6-9) shows that our method yields pronounced improvement on UT-Zappos across all three baselines, particularly when our method is integrated into Troika. In the closed-world setting, DIFF-Troika improves HM from 54.6 to 57.6 and AUC from 41.7 to 46.2. Under the more challenging open-world setting, it improves HM from 47.8 to 50.1 and AUC from 33.0 to 36.6. Consistent improvements are also observed for CSP and CAMS; for example, DIFF-CAMS increases open-world HM and AUC from 50.8 and 35.7 to 52.6 and 38.1. Given the fine-grained nature of UT-Zappos, these results suggest that diffusion-guided supervision provides complementary information for compositional recognition. Moreover, the consistent AUC gains further show that our method improves the overall trade-off between seen and unseen recognition.

\noindent\textbf{C-GQA.}
As shown in Tab.~\ref{tab:main_results} (columns 10--13), DIFFCZSL also yields consistent improvements on C-GQA, which has a larger attribute--object vocabulary and compositional search space. DIFF-Troika shows the largest improvement over its baseline, increasing closed-world HM and AUC from 29.4 and 12.4 to 32.0 and 14.3, respectively. When applied to the stronger CAMS baseline, DIFFCZSL achieves the best performance among the compared methods, obtaining an HM/AUC of 34.2/16.2 in the closed-world setting and 14.5/4.3 in the open-world setting. Although the gains over CAMS are smaller than those over Troika, their consistency across both settings shows that diffusion-guided supervision remains beneficial even when the underlying model already provides strong compositional representations. Overall, the results on C-GQA show that our method remains effective on a benchmark containing diverse real-world attribute--object concepts.

\subsection{Ablation Study}
\begin{table}[t]
\centering
\captionsetup{font=small, skip=3pt}
\setlength{\tabcolsep}{1.2pt}
\renewcommand{\arraystretch}{1.05}

\begin{minipage}[t]{0.49\columnwidth}
\centering
\caption{Ablation analysis of loss components on MIT-States and UT-Zappos under the closed-world setting.}
\label{tab:ablation1}

\resizebox{0.85\linewidth}{!}{%
\begin{tabular}{cc|cccc|cccc}
\hline
\multirow{2}{*}{$\mathcal{L}_{txt}$} &
\multirow{2}{*}{$\mathcal{L}_{img}$} &
\multicolumn{4}{c|}{MIT-States} &
\multicolumn{4}{c}{UT-Zappos} \\
\cline{3-10}
& & S$\uparrow$ & U$\uparrow$ & HM$\uparrow$ & AUC$\uparrow$
& S$\uparrow$ & U$\uparrow$ & HM$\uparrow$ & AUC$\uparrow$ \\
\hline

\xmark & \xmark
& 49.0 & 53.0 & 39.2 & 22.1
& 66.8 & 73.4 & 54.6 & 41.7 \\

\cmark & \xmark
& 49.7 & 53.3 & 39.3 & 22.4
& 71.5 & 73.6 & 57.4 & 45.4 \\

\xmark & \cmark
& 49.6 & 53.3 & 39.3 & 22.3
& 71.4 & \textbf{74.1} & 56.2 & 45.0 \\

\rowcolor{my_blue}
\cmark & \cmark
& \textbf{50.5} & \textbf{53.4}
& \textbf{39.9} & \textbf{23.1}
& \textbf{72.2} & 74.0
& \textbf{57.6} & \textbf{46.2} \\

\hline
\end{tabular}%
}
\end{minipage}
\hfill
\begin{minipage}[t]{0.49\columnwidth}
\centering
\caption{Ablation on different diffusion layers on MIT-States and UT-Zappos under the closed-world setting.}
\label{tab:ablation2}

\resizebox{0.85\linewidth}{!}{%
\begin{tabular}{c|cccc|cccc}
\hline
\multirow{2}{*}{Layer} &
\multicolumn{4}{c|}{MIT-States} &
\multicolumn{4}{c}{UT-Zappos} \\
\cline{2-9}
& S$\uparrow$ & U$\uparrow$ & HM$\uparrow$ & AUC$\uparrow$
& S$\uparrow$ & U$\uparrow$ & HM$\uparrow$ & AUC$\uparrow$ \\
\hline

Baseline
& 49.0 & 53.0 & 39.2 & 22.1
& 66.8 & 73.4 & 54.6 & 41.7 \\

Layer0
& 50.1 & 52.5 & 39.1 & 22.3
& 70.5 & \textbf{74.5} & 56.4 & 44.5 \\

Layer6
& 50.0 & 52.8 & 39.5 & 22.4
& \cellcolor{my_blue}\textbf{72.2}
& \cellcolor{my_blue}74.0
& \cellcolor{my_blue}\textbf{57.6}
& \cellcolor{my_blue}\textbf{46.2} \\

Layer9
& \cellcolor{my_blue}\textbf{50.5}
& \cellcolor{my_blue}\textbf{53.4}
& \cellcolor{my_blue}\textbf{39.9}
& \cellcolor{my_blue}\textbf{23.1}
& 71.9 & 74.2 & 57.6 & 45.7 \\

\hline
\end{tabular}%
}
\end{minipage}
\end{table}
\begin{table}[t]
\centering
\captionsetup{font=small, skip=3pt}
\renewcommand{\arraystretch}{1.05}

\begin{minipage}[t]{0.49\columnwidth}
\centering
\caption{Ablation on diffusion backbone choice (SD1.5 vs.\ SD2.1)
on MIT-States and UT-Zappos.}
\label{tab:ablation3}

\setlength{\tabcolsep}{1.2pt}

\resizebox{\linewidth}{!}{%
\begin{tabular}{c|cccc|cccc}
\hline
\multirow{2}{*}{Backbone} &
\multicolumn{4}{c|}{MIT-States} &
\multicolumn{4}{c}{UT-Zappos} \\
\cline{2-9}
& S$\uparrow$ & U$\uparrow$ & HM$\uparrow$ & AUC$\uparrow$
& S$\uparrow$ & U$\uparrow$ & HM$\uparrow$ & AUC$\uparrow$ \\
\hline

Baseline
& 49.0 & 53.0 & 39.2 & 22.1
& 66.8 & 73.4 & 54.6 & 41.7 \\

CleanDIFT (SD1.5)
& 50.1 & 53.1 & 39.5 & 22.6
& 69.9 & 73.6 & 56.3 & 43.9 \\

\rowcolor{my_blue}
CleanDIFT (SD2.1)
& \textbf{50.5}
& \textbf{53.4}
& \textbf{39.9}
& \textbf{23.1}
& \textbf{72.2}
& \textbf{74.0}
& \textbf{57.6}
& \textbf{46.2} \\

\hline
\end{tabular}%
}
\end{minipage}
\hfill
\begin{minipage}[t]{0.49\columnwidth}
\centering
\caption{Ablation on noise-conditioned vs.\ noise-free diffusion
features on MIT-States and UT-Zappos.}
\label{tab:ablation4}

\setlength{\tabcolsep}{1.2pt}

\resizebox{\linewidth}{!}{%
\begin{tabular}{c|cccc|cccc}
\hline
\multirow{2}{*}{Backbone} &
\multicolumn{4}{c|}{MIT-States} &
\multicolumn{4}{c}{UT-Zappos} \\
\cline{2-9}
& S$\uparrow$ & U$\uparrow$ & HM$\uparrow$ & AUC$\uparrow$
& S$\uparrow$ & U$\uparrow$ & HM$\uparrow$ & AUC$\uparrow$ \\
\hline

Baseline
& 49.0 & 53.0 & 39.2 & 22.1
& 66.8 & 73.4 & 54.6 & 41.7 \\

Stable Diffusion 2.1
& 49.5 & 53.3 & 39.2 & 22.3
& 71.7 & \textbf{74.1} & 55.8 & 44.7 \\

\rowcolor{my_blue}
CleanDIFT (SD2.1)
& \textbf{50.5}
& \textbf{53.4}
& \textbf{39.9}
& \textbf{23.1}
& \textbf{72.2}
& 74.0
& \textbf{57.6}
& \textbf{46.2} \\

\hline
\end{tabular}%
}
\end{minipage}
\end{table}

For a more thorough analysis, we conduct ablation studies on MIT-States~\citep{mit} and UT-Zappos~\citep{ut}, comparing against Troika~\citep{huang2024troika}. 

\noindent\textbf{Diffusion Distillation Losses.}
To disentangle the contributions of the two diffusion-guided objectives and assess their complementarity, we ablate the image and text level distillation losses in Tab.~\ref{tab:ablation1}. Removing both losses recovers the baseline performance, confirming that the improvements originate from diffusion-guided supervision. Introducing either $\mathcal{L}_{\mathrm{img}}$ or $\mathcal{L}_{\mathrm{txt}}$ individually improves the results, showing that diffusion representations provide useful supervisory signals for both the visual and semantic branches. Combining the two objectives achieves the highest HM and AUC on both MIT-States and UT-Zappos, indicating that their contributions are complementary. More importantly, these results suggest that jointly supervising the image and compositional text embeddings allows diffusion priors to regularize both sides of the shared cross-modal space, leading to a more comprehensive transfer of composition-aware structure for the CZSL task.

\noindent\textbf{Diffusion Layer Selection.}
To verify whether the depth at which diffusion features are extracted affects CZSL performance, we compare representations from different diffusion layers in Tab.~\ref{tab:ablation2}. All tested layers improve AUC over the baseline on both datasets, although their effects on the other metrics vary. On MIT-States, features from layer 9 achieve the best HM and AUC. On UT-Zappos, layer 6 achieves the highest AUC and HM. More generally, features extracted from middle-to-late layers consistently provide stronger overall performance than early-layer features across both datasets. This trend suggests that middle-to-late diffusion representations offer a more suitable balance between semantic abstraction and the preservation of compositional details. Therefore, intermediate-to-deep diffusion features generally serve as more effective and robust compositional priors for the CZSL task.

\noindent\textbf{Diffusion Backbone Choice.}
To investigate whether a more recent diffusion backbone provides semantically richer intermediate representations for supervising CZSL, we compare CleanDIFT features extracted from Stable Diffusion 1.5 and Stable Diffusion 2.1 in Tab.~\ref{tab:ablation3}. Both variants outperform the baseline across all metrics, demonstrating that diffusion-derived intermediate representations provide effective supervision regardless of the underlying backbones. Moreover, the SD2.1-based variant consistently outperforms its SD1.5-based counterpart, improving HM and AUC from 39.5 and 22.6 to 39.9 and 23.1 on MIT-States, and from 56.3 and 43.9 to 57.6 and 46.2 on UT-Zappos, respectively. Since the feature extraction and distillation framework remains unchanged, these improvements suggest that the intermediate representations of SD2.1 contain richer semantic information for compositional learning. More broadly, the results support the hypothesis that advances in diffusion pretraining can yield stronger representation-level supervision, allowing CZSL models to benefit from increasingly capable generative backbones.

\noindent\textbf{Noise Conditioning in Diffusion Feature Extraction.}
To explore whether noise injection during diffusion feature extraction affects the quality of the resulting representations, we compare noise-conditioned features from Stable Diffusion 2.1 with noise-free features from CleanDIFT in Tab.~\ref{tab:ablation4}. While both approaches extract intermediate representations from the same diffusion backbone, CleanDIFT operates directly on clean input images without introducing noise. Both variants improve the overall performance over the baseline, confirming that diffusion-derived features provide useful supervision for CZSL tasks. Compared with the noise-conditioned variant, CleanDIFT improves HM and AUC from 39.2 and 22.3 to 39.9 and 23.1 on MIT-States, and from 55.8 and 44.7 to 57.6 and 46.2 on UT-Zappos, respectively. These results suggest that injecting noise may partially influence task-relevant information and introduce variability into the extracted representations. In contrast, noise-free extraction preserves the semantic knowledge encoded by the diffusion backbone while providing more stable and informative supervisory signals.

\begin{wraptable}{r}{0.56\textwidth}
\vspace{-8pt}
\centering

\setlength{\tabcolsep}{2pt}
\renewcommand{\arraystretch}{0.92}

{\fontsize{8pt}{9pt}\selectfont
\begin{threeparttable}

\captionsetup{
    font=footnotesize,
    justification=justified,
    singlelinecheck=false,
    margin=0pt,
    skip=3pt
}
\caption{Comparison with alternative pretrained supervision on
MIT-States and UT-Zappos.}
\label{tab:test3}

\begin{tabular}{lcccccccc}
\toprule
\multirow{2}{*}{Method}
& \multicolumn{4}{c}{MIT-States}
& \multicolumn{4}{c}{UT-Zappos} \\
\cmidrule(lr){2-5}
\cmidrule(lr){6-9}
& S & U & HM & AUC
& S & U & HM & AUC \\
\midrule

Troika
& 49.0
& 53.0
& 39.2
& 22.1
& 66.8
& 73.4
& 54.6
& 41.7 \\

+ DINOv3
& 49.6
& 53.3
& 39.6
& 22.3
& 70.5
& 72.5
& 54.2
& 42.5 \\

\rowcolor{my_blue}
+ DIFF
& \textbf{50.5}
& \textbf{53.4}
& \textbf{39.9}
& \textbf{23.1}
& \textbf{72.2}
& \textbf{74.0}
& \textbf{57.6}
& \textbf{46.2} \\

\bottomrule
\end{tabular}

\end{threeparttable}
}

\vspace{-6pt}
\end{wraptable}

\noindent
{\raggedright
\textbf{Generative \textit{vs.}\ Discriminative Priors.}\par
}

\noindent
To determine whether the observed gains arise merely from adding supervision from a strong pretrained model, rather than from the semantic properties of diffusion representations, we compare our diffusion features with DINOv3 features~\citep{simeoni2025dinov3}. We select DINOv3 as the alternative supervision model because, like Stable Diffusion 2.1, it is pretrained on billion-scale image data. To ensure a controlled comparison, we keep the alignment framework and optimization settings unchanged and replace only the source of auxiliary supervision. As shown in Table~\ref{tab:test3}, DINOv3 yields only modest and inconsistent improvements. In contrast, diffusion-based supervision improves HM and AUC to 39.9 and 23.1 on MIT-States, and to 57.6 and 46.2 on UT-Zappos, respectively. Given the comparable scale of their training dataset and the controlled alignment framework, the stronger gains obtained with diffusion features are unlikely to arise solely from the introduction of a large pretrained teacher. Instead, these results suggest that diffusion representations encode semantic structures that are more relevant to attribute--object composition, making them particularly effective supervisory priors for CZSL.

\begin{figure}[!t]
    \centering
    \includegraphics[width=\linewidth]{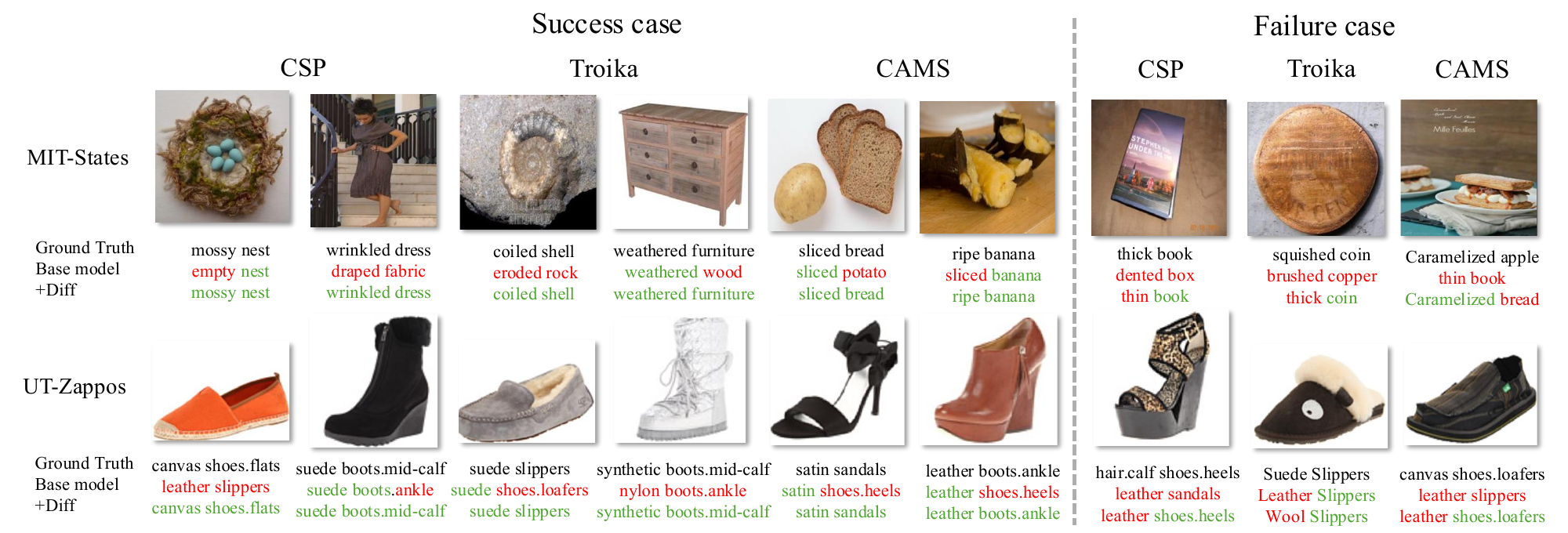}
    \caption{Visual comparison between CSP~\citep{csp},
    Troika~\citep{huang2024troika}, CAMS~\citep{cams}, and our method
    on the MIT-States~\citep{mit} and UT-Zappos~\citep{ut} datasets.
    Green denotes correct predictions, while red denotes incorrect
    predictions.}
    \label{fig:Qualitative_1}
\end{figure}

\subsection{Qualitative Analysis}
We present representative qualitative comparisons in Fig.~\ref{fig:Qualitative_1} to better understand the effect of diffusion-guided regularization. Across multiple examples, DIFFCZSL produces more semantically consistent attribute–object compositions than the CLIP-based baselines, demonstrating clearer alignment between predicted attributes and object identities. For instance, in cases such as \emph{“weathered furniture”} and \emph{“ripe banana”}, our model accurately captures both attribute and object concepts, while baseline methods occasionally confuse either the attribute or the object category.

In challenging cases involving subtle attribute variations or similar texture to background, baseline models rely on superficial visual similarity. For example, in the case of \emph{“coiled shell”}, Troika wrongly predicts \emph{“eroded rock”} due to shared texture patterns and shape ambiguity. In contrast, our method correctly identifies the compositional structure by jointly recognizing the geometric property (“coiled”) and the object identity (“shell”). This suggests that diffusion features provide stronger primitive-level meaning, enabling better generalization between state and object representations.

Nevertheless, failure cases persist. Notably, even when incorrect, our predictions are often semantically closer to the ground truth than those of CLIP-based baselines. In MIT-States, our model predicts \emph{“thin book”} for \emph{“thick book”}, preserving the object identity and a closely related attribute, whereas the baseline produces distant outputs such as \emph{“dented box”}. In UT-Zappos, for the ground-truth composition \emph{“canvas shoes.loafers”}, the baseline predicts \emph{“leather slippers”}, confusing both the material attribute and the object category. In contrast, our model predicts \emph{“leather shoes.loafers”}, successfully recovering the correct object identity while making an error only in the material attribute. This behavior suggests that diffusion-guided features favor semantically closer compositions even when the exact target is not correctly identified, indicating a more structured organization of compositional representations.

\subsection{Training Efficiency and Resource Consumption}
\label{sec:efficiency}
We evaluate the additional computational overhead of DIFFCZSL in terms of per-epoch training time and the number of trainable parameters. Because diffusion representations are used only to provide regularization during training, the diffusion feature extractor and distillation objectives are not required during inference.

\begin{wraptable}{r}{0.50\textwidth}
\vspace{-8pt}
\centering

\captionsetup{
    font=footnotesize,
    width=0.90\linewidth,
    justification=justified,
    singlelinecheck=false,
    skip=3pt
}
\caption{Training time per epoch of the three baselines and their
DIFF-augmented variants.}
\label{tab:training_time}

{\fontsize{8.5pt}{9.5pt}\selectfont
\setlength{\tabcolsep}{2pt}
\renewcommand{\arraystretch}{1.00}

\begin{tabular*}{0.90\linewidth}{
    @{\extracolsep{\fill}}lccc@{}
}
\toprule
Method & MIT-States & UT-Zappos & C-GQA \\
\midrule

CSP
& 15m 25s
& 9m 40s
& 36m 20s \\

+ DIFF
& 15m 57s
& 10m 07s
& 37m 17s \\

Troika
& 59m 26s
& 19m 36s
& 199m 35s \\

+ DIFF
& 65m 35s
& 22m 24s
& 203m 06s \\

CAMS
& 7m 35s
& 6m 36s
& 23m 39s \\

+ DIFF
& 9m 18s
& 7m 41s
& 26m 15s \\

\bottomrule
\end{tabular*}
}

\vspace{7pt}

\captionsetup{
    font=footnotesize,
    width=0.90\linewidth,
    justification=justified,
    singlelinecheck=false,
    skip=3pt
}
\caption{Additional trainable parameters introduced by DIFF.}
\label{tab:param}

{\fontsize{8.5pt}{9.5pt}\selectfont
\setlength{\tabcolsep}{2pt}
\renewcommand{\arraystretch}{1.15}

\begin{tabular*}{0.90\linewidth}{
    @{\extracolsep{\fill}}lccc@{}
}
\toprule
Method & CSP & Troika & CAMS \\
\midrule

Base model
& 27.6M
& 29.1M
& 62.1M \\

+ DIFF
& +2.7M
& +2.7M
& +17.7M \\

\bottomrule
\end{tabular*}
}

\vspace{-6pt}
\end{wraptable}

\noindent\textbf{Training Time.}
Table~\ref{tab:training_time} reports the average training time per epoch. For CSP, introducing DIFFCZSL increases the training time by only 32 seconds in MIT-States, 27 seconds in UT-Zappos, and 57 seconds in C-GQA, corresponding to relative increases of approximately 3.5\%, 4.7\%, and 2.6\%, respectively. For Troika, the additional cost is 6m09s in MIT-States, 2m48s in UT-Zappos, and 3m31s in C-GQA. Similar training overhead is observed for CAMS as well. Moreover, this additional computation is confined to training and does not affect inference efficiency.

\noindent\textbf{Parameter Overhead.}
Table~\ref{tab:param} reports the additional trainable parameters introduced by DIFFCZSL. Our method adds 2.7M parameters to CSP and Troika, increasing their model sizes from 27.6M to 30.3M and from 29.1M to 31.8M, respectively. For CAMS, DIFFCZSL adds 17.7M parameters, increasing the total parameter count from 62.1M to 79.8M. The larger increase reflects the multi-branch design of CAMS. Overall, DIFFCZSL introduces architecture-dependent training and parameter overhead while preserving the inference-time computational cost of the underlying CZSL models.

\section{Conclusion}
In this work, we study compositional zero-shot learning from the perspective of integrating generative priors into discriminative vision-language models. We find that compositional generalization can be improved by incorporating complementary semantic structure from diffusion models. To this end, we propose DIFFCZSL, a simple plug-and-play framework that introduces diffusion representations as auxiliary priors to regularize CLIP-based CZSL models. By extracting intermediate features from pre-trained diffusion models and aligning them with visual and textual representations, our method encourages the embedding space to capture more structured, composition-aware semantics, while preserving the original inference pipeline. Extensive experiments on MIT-States, UT-Zappos50K, and C-GQA demonstrate that our approach consistently improves strong CLIP-based baselines under both closed- and open-world settings. These results highlight that diffusion-derived features provide complementary supervision from visual and semantic perspectives, leading to more robust compositional representations. Overall, our findings suggest that diffusion representations offer a promising direction for enhancing discriminative vision-language models in compositional reasoning tasks, with minimal additional computational and model complexity. We hope this work inspires future research on integrating generative priors to further bridge the gap between generative and discriminative paradigms for improved generalization.


\bibliography{main}
\bibliographystyle{tmlr}

\appendix
\section{Appendix}
\subsection{Experiment Details}

\subsubsection{Implementation Details and Hyperparameters}
\label{app:implementation}
\begin{table}[t]
\centering
\caption{Implementation details and hyperparameters for DIFF-CSP across different datasets.}
\setlength{\tabcolsep}{3pt}
\label{tab:csp}
\begin{tabular}{lccc}
\toprule
\textbf{Hyperparameter} & \textbf{MIT-States} & \textbf{UT-Zappos} & \textbf{C-GQA} \\

\midrule
Learning rate & $5 \times 10^{-5}$ & $5 \times 10^{-4}$ & $5 \times 10^{-5}$ \\
Batch size & 64 & 32 & 16 \\
Gradient accumulation & 2 & 2 & 4 \\
Epochs & 20 & 20 & 15 \\

Image alignment weight $\lambda_{\text{img}}$ & 0.1 & 0.5 & 0.1 \\
Text alignment weight $\lambda_{\text{txt}}$ & 1 & 0.5 & 0.8 \\
Temperature $\tau$ & 0.07 & 0.07 & 0.07 \\

External feature dim & 640 & 1280 & 1280 \\
Projector hidden dim & 768 & 1536 & 1536 \\
Projector learning rate & $5 \times 10^{-5}$ & $5 \times 10^{-4}$ & $5 \times 10^{-5}$ \\

Diffusion feature layer & layer 9 & layer 0 & layer 0 \\
\bottomrule
\end{tabular}
\end{table}
\begin{table}[t]
\centering
\caption{Implementation details and hyperparameters for  DIFF-Troika across different datasets.}
\setlength{\tabcolsep}{3pt}
\label{tab:troika}
\begin{tabular}{lccc}

\toprule
\textbf{Hyperparameter} & \textbf{MIT-States} & \textbf{UT-Zappos} & \textbf{C-GQA} \\
\midrule
Learning rate & $1 \times 10^{-4}$ & $5 \times 10^{-4}$ & $5 \times 10^{-5}$ \\
Batch size & 8 & 16 & 16 \\
Gradient accumulation & 8 & 4 & 1 \\
Epochs & 10 & 10 & 15 \\

Image alignment weight $\lambda_{\text{img}}$ & 0.7 & 0.5 & 0.7 \\
Text alignment weight $\lambda_{\text{txt}}$ & 0.3 & 0.5 & 0.3 \\
Temperature $\tau$ & 0.1 & 0.1 & 0.1 \\

External feature dim & 640 & 1280 & 640 \\
Projector hidden dim & 768 & 1536 & 768 \\
Projector learning rate & $1 \times 10^{-4}$ & $5 \times 10^{-5}$ & $5 \times 10^{-5}$ \\

Diffusion feature layer & layer 9 & layer 6 & layer 9 \\
\bottomrule
\end{tabular}
\end{table}
\begin{table}[t]
\centering
\caption{Implementation details and hyperparameters for the DIFF-CAMS across different datasets.}
\setlength{\tabcolsep}{3pt}
\label{tab:cams}
\begin{tabular}{lccc}
\toprule
\textbf{Hyperparameter} & \textbf{MIT-States} & \textbf{UT-Zappos} & \textbf{C-GQA} \\
\midrule
Learning rate & $1 \times 10^{-4}$ & $2.5 \times 10^{-4}$ & $1 \times 10^{-4}$ \\
Batch size & 64 & 64 & 16 \\
Gradient accumulation & 1 & 1 & 1 \\
Epochs & 15 & 15 & 15 \\

Image alignment weight $\lambda_{\text{img}}$ & 0.5 & 0.5 & 0.5 \\
Text alignment weight $\lambda_{\text{txt}}$ & 0.1 & 0.1 & 0.5 \\
Temperature $\tau$ & 0.1 & 0.1 & 0.1 \\

External feature dim & 1280 & 1280 & 1280 \\
Projector hidden dim & 1536 & 1536 & 1536 \\
Projector learning rate & $1 \times 10^{-4}$ & $5 \times 10^{-5}$ & $5 \times 10^{-5}$ \\

Diffusion feature layer & layer 0 & layer 6 & layer 0 \\
\bottomrule
\end{tabular}
\end{table}
We implement our framework in PyTorch based on three representative CLIP-based CZSL baselines, namely CSP~\citep{csp}, Troika~\citep{huang2024troika}, and CAMS~\citep{cams}. We adopt the pre-trained CLIP ViT-L/14 model~\citep{radford2021learning} as the image and text encoder, and use CleanDIFT~\citep{stracke2025cleandift} built upon Stable Diffusion 2.1~\citep{rombach2022high} to extract diffusion features. During feature extraction, we use simple textual prompts corresponding to primitive concepts (\texttt{[attribute]} \texttt{[object]}). The extracted diffusion features are globally average pooled to obtain compact features.

To bridge the representation gap between diffusion and CLIP embedding spaces, we employ two lightweight projection heads for the visual and textual branches, respectively. Each projection head is implemented as a lightweight two-layer MLP with a ReLU activation, mapping diffusion features into the corresponding CLIP embedding space.

The implementation details and hyperparameters vary across baselines and datasets. We report the full configuration for DIFF-CSP, DIFF-Troika, and DIFF-CAMS on MIT-States~\citep{mit}, UT-Zappos~\citep{ut}, and C-GQA~\citep{cgqa} in Tabs.~\ref{tab:csp}-\ref{tab:cams}. These hyperparameters include the learning rate, batch size, gradient accumulation steps, number of epochs, alignment weights, temperature, external feature dimension, hidden projector dimension, projector learning rate, and selected diffusion feature layer.

\subsection{Additional Analysis}

All analyses in this section are conducted based on our DIFFCZSL framework built upon Troika, evaluated on MIT-States and UT-Zappos under the closed-world setting, unless otherwise specified.
\subsubsection{Sensitivity to Alignment Loss Weights}

\begin{table}[t]
\centering

\begin{minipage}[t]{0.49\linewidth}
\vspace{0pt}
\centering

\captionsetup{
    font=normalsize,
    justification=justified,
    singlelinecheck=false,
    margin=0pt,
    width=\linewidth,
    skip=3pt
}
\captionof{table}{
Sensitivity to $\lambda_{\mathrm{img}}$ on UT-Zappos under the closed-world setting.
}
\label{tab:img}

{\fontsize{9.5pt}{10.5pt}\selectfont
\renewcommand{\arraystretch}{1.05}
\begin{tabularx}{\linewidth}{@{}*{6}{Y}@{}}
\toprule
$\lambda_{\mathrm{img}}$
& $\lambda_{\mathrm{txt}}$
& Seen
& Unseen
& HM
& AUC \\
\midrule

0.1
& 0.5
& 71.6
& 73.8
& 56.8
& 45.1 \\

0.3
& 0.5
& 71.3
& 73.9
& 56.8
& 45.4 \\

\rowcolor{my_blue}
0.5
& 0.5
& \textbf{72.2}
& 74.0
& \textbf{57.6}
& \textbf{46.2} \\

1.0
& 0.5
& 70.8
& \textbf{74.7}
& 56.2
& 44.8 \\

\bottomrule
\end{tabularx}
}

\end{minipage}
\hfill
\begin{minipage}[t]{0.49\linewidth}
\vspace{0pt}
\centering

\captionsetup{
    font=normalsize,
    justification=justified,
    singlelinecheck=false,
    margin=0pt,
    width=\linewidth,
    skip=3pt
}
\captionof{table}{
Sensitivity to $\lambda_{\mathrm{txt}}$ on UT-Zappos under the closed-world setting.
}
\label{tab:txt}

{\fontsize{9.5pt}{10.5pt}\selectfont
\renewcommand{\arraystretch}{1.05}
\begin{tabularx}{\linewidth}{@{}*{6}{Y}@{}}
\toprule
$\lambda_{\mathrm{img}}$
& $\lambda_{\mathrm{txt}}$
& Seen
& Unseen
& HM
& AUC \\
\midrule

0.5
& 0.1
& 71.6
& \textbf{74.9}
& \textbf{57.6}
& 46.0 \\

0.5
& 0.3
& 71.7
& 74.4
& 57.1
& 45.7 \\

\rowcolor{my_blue}
0.5
& 0.5
& \textbf{72.2}
& 74.0
& \textbf{57.6}
& \textbf{46.2} \\

0.5
& 1.0
& 71.4
& 74.6
& 57.4
& 45.8 \\

\bottomrule
\end{tabularx}
}

\end{minipage}

\end{table}

We analyze the sensitivity of the model to the alignment weights $\lambda_{\text{img}}$ and $\lambda_{\text{txt}}$ by varying one factor at a time. Specifically, we first vary $\lambda_{\text{img}} \in \{0.1, 0.3, 0.5, 1.0\}$ while fixing $\lambda_{\text{txt}} = 0.5$, and then vary $\lambda_{\text{txt}} \in \{0.1, 0.3, 0.5, 1.0\}$ while fixing $\lambda_{\text{img}} = 0.5$. From Table~\ref{tab:img}, performance improves as $\lambda_{\text{img}}$ increases to a moderate value, after which it slightly degrades, suggesting that overly strong visual alignment may hurt generalization. From Table~\ref{tab:txt}, the model is relatively stable across different $\lambda_{\text{txt}}$ values, with the best performance achieved at $\lambda_{\text{txt}} = 0.5$. Both smaller and larger values lead to slight performance drops, indicating that an appropriate choice of $\lambda_{\text{txt}}$ is important to achieve good performance. Overall, the best performance is achieved at $\lambda_{\text{img}}=0.5$ and $\lambda_{\text{txt}}=0.5$.

\subsubsection{Projector Architecture and Design Choices}
\begin{table}[t]
\centering

\begin{minipage}[t]{0.49\linewidth}
\vspace{0pt}
\centering

\captionsetup{
    font=normalsize,
    justification=justified,
    singlelinecheck=false,
    margin=0pt,
    width=0.96\linewidth,
    skip=3pt
}
\captionof{table}{
Effect of projector architecture on UT-Zappos.}
\label{tab:projector}

{\fontsize{9.5pt}{10.5pt}\selectfont
\renewcommand{\arraystretch}{1.05}
\begin{tabularx}{0.96\linewidth}{@{}l*{4}{Y}@{}}
\toprule
Architecture & Seen & Unseen & HM & AUC \\
\midrule

1-layer (Linear)
& 70.8
& 73.1
& 55.9
& 44.8 \\

\rowcolor{my_blue}
2-layer (ReLU)
& \textbf{72.2}
& \textbf{74.0}
& \textbf{57.6}
& \textbf{46.2} \\

3-layer (ReLU)
& 71.7
& 73.8
& 57.0
& 45.6 \\

2-layer (GELU)
& 71.5
& 73.6
& 56.8
& 45.3 \\

\bottomrule
\end{tabularx}
}

\end{minipage}
\hfill
\begin{minipage}[t]{0.49\linewidth}
\vspace{0pt}
\centering

\captionsetup{
    font=normalsize,
    justification=justified,
    singlelinecheck=false,
    margin=0pt,
    width=0.96\linewidth,
    skip=3pt
}
\captionof{table}{
Effect of diffusion prompt templates on UT-Zappos.
}
\label{tab:prompt_ablation}

{\fontsize{9.5pt}{10.5pt}\selectfont
\renewcommand{\arraystretch}{1.05}
\begin{tabularx}{0.96\linewidth}{@{}l*{4}{Y}@{}}
\toprule
Prompt & Seen & Unseen & HM & AUC \\
\midrule

\rowcolor{my_blue}
Prompt 1
& \textbf{72.2}
& 74.0
& \textbf{57.6}
& \textbf{46.2} \\

Prompt 2
& 71.5
& \textbf{74.5}
& 57.2
& 45.8 \\

Prompt 3
& 70.5
& 73.4
& 56.4
& 44.5 \\

\bottomrule
\end{tabularx}
}

\vspace{3pt}

\begin{minipage}{0.96\linewidth}
{\fontsize{7.5pt}{8.5pt}\selectfont
\raggedright
\textit{Prompt 1}: ``[attribute] [object]'';\quad
\textit{Prompt 2}: ``a photo of [attribute] [object]'';\quad
\textit{Prompt 3}: ``a [object] that is [attribute]''.
\par}
\end{minipage}

\end{minipage}

\end{table}
As shown in Table~\ref{tab:projector}, we analyze the impact of the projection head design from two perspectives: network depth and activation function.

\noindent\textbf{Projector Depth Analysis.}
We compare projection heads with different depths, including a single-layer linear projection, a two-layer MLP (our default setting), and a three-layer MLP. The results show that the two-layer MLP achieves the best performance, while a single-layer projection lacks sufficient capacity to bridge the representation gap, and a deeper three-layer MLP does not bring further improvement.

\noindent\textbf{Activation Function Analysis.}
We further compare different activation functions in the two-layer MLP, including ReLU (default) and GELU. We observe that ReLU performs slightly better and is more stable in our setting, possibly because the projection task does not require highly complex non-linear transformations.

Overall, these results suggest that a lightweight two-layer MLP with ReLU provides a good balance between model capacity and training stability for aligning diffusion and CLIP representations.

\subsubsection{Analysis of Prompt Design on Compositional Representations}
\begin{table*}[t]
\centering
\caption{Performance comparison between Troika and our method on MIT-States and UT-Zappos, averaged over 3 random seeds with standard deviations. † marks results taken from the original publications.}
\label{tab:supple_ablation4}

\setlength{\tabcolsep}{3pt}
\renewcommand{\arraystretch}{1.1}

\begin{tabular}{l|cccc|cccc}
\hline
\multirow{2}{*}{Method} &
\multicolumn{4}{c|}{MIT-States} &
\multicolumn{4}{c}{UT-Zappos} \\
\cline{2-9}
& S & U & HM & AUC
  & S & U & HM & AUC \\
\hline

Troika† (baseline)
& 49.0{\scriptsize$\pm$0.4} 
& 53.0{\scriptsize$\pm$0.2}
& 39.3{\scriptsize$\pm$0.2} 
& 22.1{\scriptsize$\pm$0.1}
& 66.8{\scriptsize$\pm$1.1} 
& 73.8{\scriptsize$\pm$0.6} 
& 54.6{\scriptsize$\pm$0.5} 
& 41.7{\scriptsize$\pm$0.7} \\

\rowcolor{my_blue}
\textbf{DIFF-Troika (ours)}
& \textbf{49.9{\scriptsize$\pm$0.6}} 
& \textbf{53.2{\scriptsize$\pm$0.2}} 
& \textbf{39.6{\scriptsize$\pm$0.3}} 
& \textbf{22.6{\scriptsize$\pm$0.4}}
& \textbf{71.0{\scriptsize$\pm$1.2}} 
& \textbf{74.4{\scriptsize$\pm$0.4}} 
& \textbf{57.7{\scriptsize$\pm$0.2}} 
& \textbf{45.4{\scriptsize$\pm$0.8}} \\

\hline
\end{tabular}

\end{table*}
We study the effect of different prompt templates used to extract external features from the diffusion model. As shown in Table~\ref{tab:prompt_ablation}, prompt 1 achieves the best overall performance, while prompt 2 gives a slightly higher unseen accuracy. Overall, the simple template in prompt 1 works better than the more natural-language formulations in prompt 2 and prompt 3. This suggests that, for our setting, a concise compositional prompt provides more effective supervision than longer sentence-style templates when constructing diffusion-based representations. These results indicate that the choice of prompt template has a non-negligible effect on downstream CZSL performance.

\subsubsection{Stability Analysis}

Following the standard practice~\citep{huang2024troika}, we evaluate the stability of our method under 3 different random seeds. As shown in Table~\ref{tab:supple_ablation4}, the performance remains consistent across multiple runs, with only minor variations in all metrics. The standard deviation is relatively small, indicating that the performance gains are stable and not sensitive to random initialization. These results demonstrate that our method is robust and that the improvements are not due to favorable random seeds.

\subsection{Relational Analysis of CLIP and Diffusion Feature Spaces}
\begin{wraptable}{r}{0.59\textwidth}
\vspace{-8pt}
\centering

\captionsetup{
    font=normalsize,
    justification=justified,
    singlelinecheck=false,
    margin=0pt,
    width=\linewidth,
    skip=3pt
}
\caption{
Summary statistics of per-attribute consistency for diffusion and CLIP feature spaces on MIT-States.}
\label{tab:attr_consistency}
{\fontsize{8.5pt}{9.5pt}\selectfont
\renewcommand{\arraystretch}{1.05}
\begin{tabularx}{
    \linewidth
}{
    @{}
    l
    *{5}{>{\centering\arraybackslash}X}
    @{}
}
\toprule
Method
& Mean $\uparrow$
& Median $\uparrow$
& Q25 $\uparrow$
& Q75 $\uparrow$
& Max $\uparrow$ \\
\midrule

CLIP
& 0.1887
& 0.1818
& 0.1342
& 0.2310
& 0.4561 \\

\rowcolor{my_blue}
Diffusion
& \textbf{0.2012}
& \textbf{0.1935}
& \textbf{0.1552}
& \textbf{0.2456}
& \textbf{0.5654} \\

\bottomrule
\end{tabularx}
}

\vspace{-6pt}
\end{wraptable}

We evaluate relational compositionality by measuring the consistency of attribute-induced displacement vectors across different object contexts. Concretely, for each attribute--object pair $(a,o)$, we compute a class centroid by averaging the image features of all samples belonging to that composition, followed by $\ell_2$ normalization. This is performed separately for diffusion and CLIP features, yielding one centroid vector $f(a,o)$ for each composition in each feature space.

Next, for a pair of attributes $(a_1,a_2)$, we collect all objects $o$ that appear with both attributes. For each shared object, we compute the attribute-induced displacement vector
\[
d_o = f(a_1,o) - f(a_2,o).
\]
If a representation space captures attributes in a context-consistent way, then changing from $a_2$ to $a_1$ should induce similar displacement directions even when the object changes. Therefore, for each attribute pair, we measure consistency as the mean pairwise cosine similarity among all displacement vectors $\{d_o\}$ computed from different shared objects.

This formulation is inspired by prior work on vector-space embeddings, where semantic relations can be approximated by linear transformations in the embedding space~\citep{mikolov2013linguistic}~\citep{mikolov2013efficient}. Under this perspective, compositionality can be partially reflected by whether the same semantic change induces similar displacement directions across contexts.

As shown in Table~\ref{tab:attr_consistency}, diffusion features achieve higher mean and median consistency compared to CLIP, indicating that attribute-induced transformations are more stable across object contexts. This suggests that diffusion features provide more consistent attribute-related transformations across different objects. Such a property is desirable for compositional generalization, where attributes must be consistently applied to novel attribute and object combinations.

\end{document}